\documentclass[conference]{IEEEtran}
\usepackage{times}

\usepackage[numbers]{natbib}
\usepackage{multicol}
\usepackage[bookmarks=true]{hyperref}

\usepackage{fancyhdr}
\fancypagestyle{firstpage}{%
  \chead{\color{gray} \large This paper has been accepted for publication at Robotics: Science and Systems, 2022.}
}
\usepackage{mathtools}
\usepackage{amssymb}
\usepackage{amsmath}
\usepackage{dblfloatfix} 
\usepackage{comment}
\usepackage{hyperref}
\usepackage{graphicx}
\usepackage{subcaption}
\usepackage{caption}
\usepackage{xcolor}
\usepackage{siunitx}
\usepackage[section]{placeins}
\usepackage[ruled,vlined,linesnumbered]{algorithm2e}
\DeclareMathOperator*{\argmax}{arg\,max}

\newcommand{\norm}[1]{\left\lVert#1\right\rVert}
\newcommand{\ie}{{i.e.}}
\newcommand{\eg}{{e.g.}}
\newcommand{\realR}{\mathbb{R}}
\newcommand{\circlegroup}{\mathbb{T}}

\title{
Cooperative Multi-Agent Trajectory Generation \\ with Modular Bayesian Optimization
}

\author{\authorblockN{Gilhyun~Ryou, Ezra~Tal, and~Sertac~Karaman}
\authorblockA{
Massachusetts Institute of Technology, \\
Cambridge, Massachusetts 02139\\
Email: $\{$ghryou,eatal,sertac$\}$@mit.edu}}

\begin{document}

\maketitle
\thispagestyle{firstpage}

\begin{abstract}
We present a modular Bayesian optimization framework that efficiently generates time-optimal trajectories for a cooperative multi-agent system, such as a team of UAVs.
Existing methods for multi-agent trajectory generation often rely on overly conservative constraints to reduce the complexity of this high-dimensional planning problem, leading to suboptimal solutions.
We propose a novel modular structure for the Bayesian optimization model that consists of multiple Gaussian process surrogate models that represent the dynamic feasibility and collision avoidance constraints.
This modular structure alleviates the stark increase in computational cost with problem dimensionality and enables the use of minimal constraints in the joint optimization of the multi-agent trajectories.
The efficiency of the algorithm is further improved by introducing a scheme for simultaneous evaluation of the Bayesian optimization acquisition function and random sampling.
The modular BayesOpt algorithm was applied to optimize multi-agent trajectories through six unique environments using multi-fidelity evaluations from various data sources.
It was found that the resulting trajectories are faster than those obtained from two baseline methods.
The optimized trajectories were validated in real-world experiments using four quadcopters that fly within centimeters of each other at speeds up to 7.4 m/s.
\end{abstract}

\IEEEpeerreviewmaketitle

\section*{Supplementary Material}
A video of the experiments is available at \url{https://youtu.be/rxQiNeXvLTc}.
Results and demonstrations also can be found on our project website: \url{https://aera.mit.edu/projects/MultiDroneModularBayesOpt}.
\vspace{1em}
\section{Introduction} \label{sections:introduction}

In this paper, we study a multi-agent trajectory optimization problem in which quadcopter vehicles are tasked with traversing a complex environment as fast as possible while avoiding collisions with obstacles and with each other.
We consider the problem of designing trajectories for all vehicles involved, minimizing the total time of the mission.
Specifically, we are motivated by robotics applications in which multiple vehicles must simultaneously visit certain locations, \eg, to collect synchronized sensor measurements from different viewpoints or to complete a coordinated task or motion in a cooperative manner.
These applications add the requirement for agents to pass through multi-agent formation waypoints synchronously.

%
%

The problem is an instance of \textit{cooperative} multi-agent planning, in contrast to \textit{non-cooperative} scenarios in which agents have opposing objectives, such as in multi-robot racing~\cite{wang2019game}.
Existing literature has considered cooperative multi-agent motion planning in various contexts, including for unmanned aerial vehicle (UAV) systems.
%
In multi-agent sensor systems, motion planning can be applied to cooperatively gather and share data~\cite{schwager2009optimal, tian2020search}.
Multi-agent systems may also cooperate against a shared adversary, \eg, in target or perimeter defense games where a team of UAVs aims to stop intruders~\cite{liang2019differential, shishika2020cooperative}.

These multi-agent planning problems have two properties in common that are particularly relevant when trajectories must be as fast as possible.
First, collision avoidance between agents should be considered in a spatio-temporal manner, which means that trajectories may intersect as long as vehicles pass through the intersection at different times.
Second, vehicles are only required to attain their position within the multi-agent formation at specific points in the trajectory.
This implies that---when traveling between these specific waypoints---agents may deviate from the formation in order to achieve more efficient, \ie, faster, trajectories.

\begin{figure}[t!]
	\centering
	\includegraphics[width=0.48\textwidth,trim=.0cm -1.0cm .0cm -2.0cm,clip]{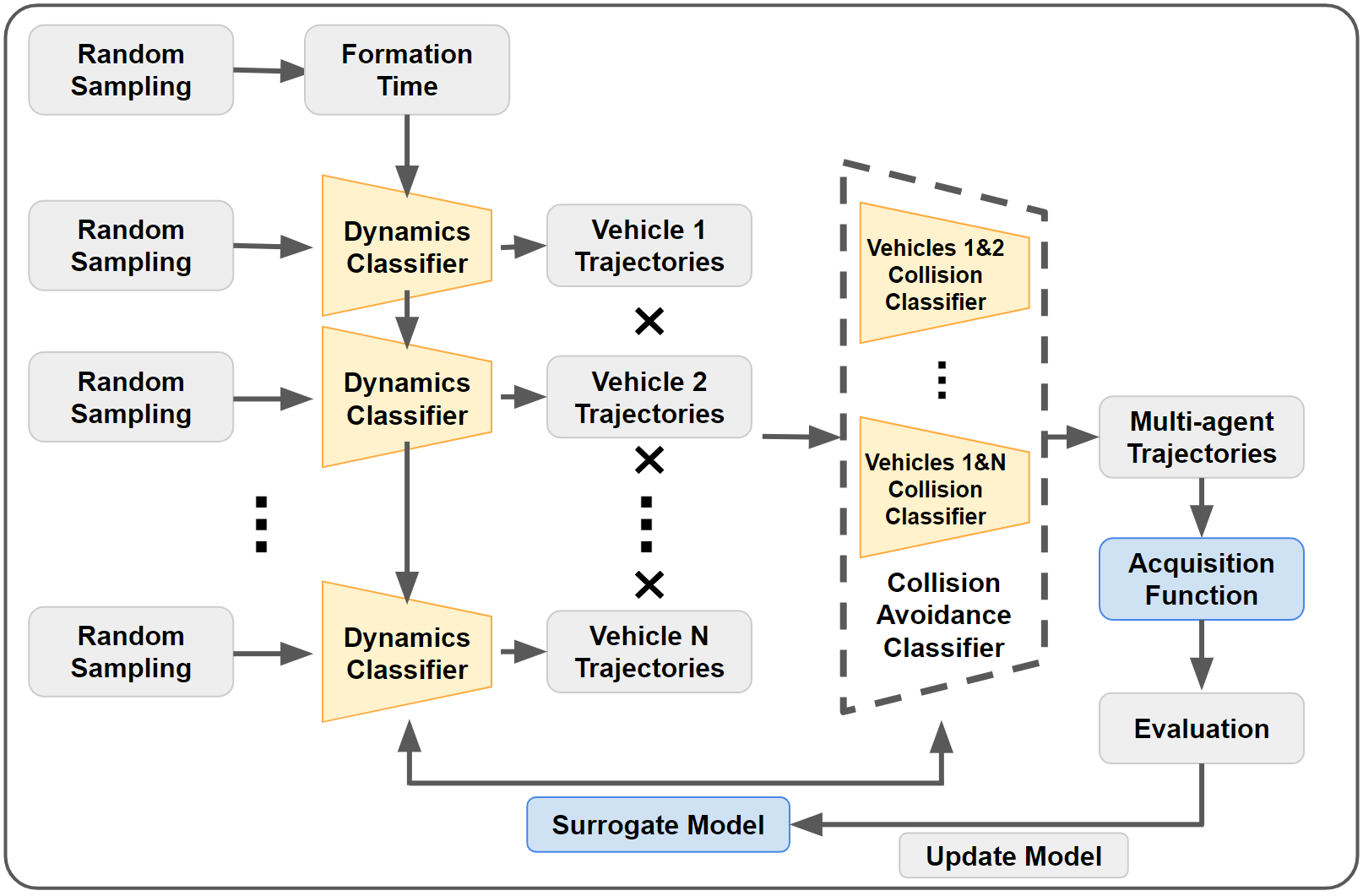}
	\caption{Overview of our proposed algorithm. The BayesOpt model is composed of multiple Gaussian process models, which represent dynamic feasibility and collision avoidance constraints.}
	\label{fig:diagram_main_algorithm}
\end{figure}

Our aim is to generate time-optimal multi-agent trajectories that connect specified start and end points and synchronously attain a sequence of formation waypoints along the way.
In order to achieve time optimality, we must explicitly leverage the two properties described above.
This is challenging because spatio-temporal collision avoidance and formation synchronization require joint and holistic consideration of the agent trajectories.
Consequently, the dimension of the input space rapidly increases with the number of agents, leading to prohibitive computational cost.
The problem is further complicated by the fact that fast and agile flight in tight formations is intrinsically complex.
Unpredictable flight dynamics and subsystem limitations (\eg, in control and actuation) necessitate more sophisticated and computationally costly methods to predict dynamic feasibility of candidate trajectories~\cite{ryou2021ijrr}.
These methods are especially needed in time-optimal multi-agent trajectories where vehicles may maneuver in close proximity of each other so that even a small deviation from the anticipated trajectory can result in a collision.

In this paper, we propose an algorithm that enables efficient optimization of multi-vehicle trajectories by alleviating the explosion of computational cost commonly associated with high dimensionality.
The algorithm leverages Bayesian optimization (BayesOpt) to efficiently sample the search space and build surrogate models that represent the dynamic feasibility and collision avoidance constraints.
While BayesOpt has been applied towards trajectory optimization~\cite{ryou2021ijrr}, we present significant innovations that enable application to the high-dimensional cooperative multi-agent planning problem.
In addition to the resulting trajectory optimization algorithm, we present several other contributions.
First, we present a modular Bayesian optimization architecture that reduces the computational burden of multi-agent optimization.
Second, we combine the BayesOpt acquisition function evaluation and random sampling steps to further improve the overall efficiency of Bayesian optimization.
Third, we demonstrate that our novel BayesOpt architecture can be applied in a multi-fidelity framework with objective and constraint evaluations from various data sources.
Fourth, we present extensive experimental results from the proposed algorithm in six unique environments with obstacles and we evaluate the resulting trajectories in both simulation and real-world flight experiments.

%

\section{Preliminaries} \label{sections:preliminaries}
\subsection{Problem Definition}
Our goal is to generate time-optimal multi-agent quadrotor trajectories that (i) connect start and end points, (ii) attain specified formation waypoints, and (iii) avoid any collision with obstacles or between vehicles.
For each vehicle, a trajectory is represented by a continuous function $p : \realR_{\geq 0} \rightarrow \realR^3 \times \circlegroup$---with $\circlegroup$ the circle group---that maps time to position and yaw, \ie, $p(t) = \begin{bmatrix}{p_r(t)}^T &  p_\psi(t)\end{bmatrix}^T$.
Along these trajectories, there are $N_f$ waypoints at which the $N_v$ vehicles must synchronously attain a specified formation.
We denote the times at which these formations are attained by $\mathbf{t}=\begin{bmatrix}
t_{1}^\text{form}&\cdots&t_{N_f}^\text{form}
\end{bmatrix}$ and the corresponding position and yaw waypoints for vehicle $i$ by $\mathbf{\tilde p}_i^\text{form}=\begin{bmatrix}\tilde{p}_{i,1}^\text{form}&\cdots&\tilde{p}_{i,N_f}^\text{form}\end{bmatrix}$.
The total trajectory time for vehicle $i$ is indicated by $T_i$, so that minimum-time multi-agent planning problem is defined as:
\vspace{1em}
\begingroup
\begin{equation}\label{eqn:planning_general}
\underset{\mathbf{p},\; \mathbf{t},\; \mathbf{T}}{\text{minimize}}
\max_{i=1,\dots,N_v} T_i
\end{equation}
\vspace{-2em}
\begin{multline*}
\text{subject to} \hfill\\
p_i(0) = \tilde{p}_i^\text{start},\;p_i(T_i) = \tilde{p}_i^\text{end},\hfill i=1,\dots,N_v, \\
p_i(t^\text{form}_{k}) = \tilde{p}_{i,k}^\text{form}, \hfill i=1,\dots,N_v,\;k=1,\dots,N_f,\\
t^\text{form}_{k}\leq t^\text{form}_{k+1},\hfill k=1,\dots,N_f-1,\\
t^\text{form}_{N_f} \leq T_i,\;p_i \in \mathcal{P}_{T_i},\;p_i \in \mathcal{F}_{T_i},\hfill i=1,\dots,N_v, \\
\left(p_i, p_j\right) \in \mathcal{F}_{T_i,T_j}, \hfill i,j=1,\dots,N_v,\;j>i,\hspace{-1em}
\end{multline*}
\endgroup
where $\tilde{p}_i^\text{start}$ and $\tilde{p}_i^\text{end}$ are respectively the start and end position and yaw of vehicle $i$,
and $\mathbf{p} = \{p_i\}_{i=1,\dots,N_v}$ and $\mathbf{T}=\begin{bmatrix}
T_{1}&\cdots&T_{N_v}
\end{bmatrix}$.
The function space $\mathcal{P}_{T_i}$ is the set of trajectories that satisfy the feasibility constraints over the time interval $[0,T_i]$, \ie, it contains all trajectory functions that the quadrotor can successfully track.
Similarly, $\mathcal{F}_{T_i}$ denotes trajectories that satisfy the obstacle avoidance constraints, and $\mathcal{F}_{T_i,T_j}$ denotes trajectory pairs that do not collide.

\subsection{Quadrotor Trajectory Generation}\label{sec:quadplan}
The quadrotor is a highly-agile platform that is capable of tracking challenging maneuvers at high speeds. 
During such maneuvers, it may exhibit highly nonlinear dynamics, complicating the integration of the feasibility constraints that define $\mathcal{P}_{T_i}$ in trajectory optimization.
Popular methods for trajectory planning avoid the complex dynamics constraints by reformulating the optimization problem such that dynamic feasibility is the objective, instead of a constraint~\cite{mellinger2011minimum,richter2016polynomial}.
In practice, this is achieved by minimizing high-order derivatives of the trajectory function, particularly, by minimizing the fourth-order derivative of position (\ie, snap) and the second-order derivative of yaw.
As these derivatives are related to the angular acceleration of the vehicle, their minimization reduces the required control moment and thereby increases the likelihood that the required motor speeds remain feasible.
This approach leads to the following objective function
\begin{equation}\label{eqn:smoothness}
\sigma(p,T)=\int_{0}^{T} \mu_r \norm{\frac{d^4 p_{r}}{dt^4}}^2 + \mu_\psi \Big(\frac{d^2 p_\psi}{dt^2}\Big)^2 dt,
\end{equation}
where $\mu_r$ and $\mu_\psi$ are weighing parameters.

Obstacle avoidance constraints can be incorporated in the trajectory optimization by using polytope constraints to describe their boundaries~\cite{deits2015efficient, gao2018online, ryou2021ijrr}.
Since polytope constraints can be described by linear inequalities, their integration into the optimization formulation does not increase its complexity.
In this paper, we utilize the obstacle constraint formulation from \cite{ryou2021ijrr}, which decomposes the obstacle-free space into convex polytopes, resulting in the following minimum-snap optimization:
\begin{equation}
\begin{aligned}
&\underset{p}{\text{minimize}}
& & \sigma(p,\sum \nolimits_{i=1}^m x_i)\\
&  \text{subject to} & & p(0) = \tilde{p}^\text{start}, \:p\left(\sum \nolimits_{i=1}^m x_i\right) = \tilde{p}^\text{end}, \\
&   & & A_i p\left(t\right) \leq b_i, \; \forall t\in\left[\sum \nolimits_{j=1}^{i-1}x_j, \sum \nolimits_{j=1}^{i}x_j\right],\\
&& &\hspace{6em} i=1,\;\dots,\;m,\\
\label{eqn:minsnap}
\end{aligned}
\end{equation}
where the matrix $A_i \in \mathbb{R}^{d_i \times 3}$ and the vector $b_i \in \mathbb{R}^{d_i}$ constrain the $i$-th trajectory segment to be within a polytope of $d_i$ faces.
The vector $\mathbf{x} = \begin{bmatrix}{x}_1&\cdots&{x}_{m}\end{bmatrix}$ contains the time allocation over the trajectory segments corresponding to these polytopes. 
By using a piecewise polynomial representation of $p$, we can effectively represent the set of trajectories that attains these geometric constraints.

By combining the polytope and formation waypoint constraints, we can describe a multi-agent trajectory through $N_{p}$ polytopes with $N_f$ formation waypoints using $m = N_{p} + N_f$ trajectory segments per vehicle.
The time allocation for a single vehicle can be written as $\mathbf{x} = \begin{bmatrix}x_{e_{0}}&\cdots&x_{e_{k}}&\cdots&x_{e_{N_f+1}}\end{bmatrix}$, where $e_0=1$, $e_{N_f+1}=m$, and $e_k$ is the index of the trajectory segment that ends at the $k$-th formation waypoint, \ie,
\begin{equation}\label{eq:wpconstraint}
p(\sum_{i=1}^{e_{k}} x_{i}) = \tilde{p}^\text{form}_{k}.
\end{equation}
For convenience, we denote the function that gives the minimizer trajectory of \eqref{eqn:minsnap} with \eqref{eq:wpconstraint} for a given time allocation $\mathbf{x}$ as follows:
\begin{equation}\label{eqn:min_snap_smoothness}
p = \chi(\mathbf{x}, \tilde{\mathcal{F}}),
\end{equation}
where $\tilde{\mathcal{F}}$ represents $(\tilde{p}^\text{start}, \tilde{p}^\text{end}, \mathbf{A}, \mathbf{b}, \mathbf{\tilde{p}^\text{form}})$ with $\mathbf{A}$ and $\mathbf{b}$ containing respectively all $A_i$ and $b_i$.

Minimum-snap trajectory generation algorithms commonly employ a two-step process based on \eqref{eqn:min_snap_smoothness}. First, the minimum-snap trajectory for a (very large) initial guess of the total trajectory time is found, as follows:
\begin{equation}
\begin{aligned}
&\underset{\mathbf{x}\in \realR^m_{\geq 0}}{\text{minimize}}
& & \sigma\left(\chi(\mathbf{x},\tilde {\mathcal{F}}),T\right)\\
&  \text{subject to} & & T =\sum \nolimits_{i=1}^m x_i.
\label{eqn:minsnap-2}
\end{aligned}
\end{equation}
Next, the obtained time allocation is scaled down to obtain the minimum-time trajectory, \ie,
\begin{equation}
\begin{aligned}
&\underset{\eta\in\realR_{> 0}}{\text{minimize}}
& & T\\
&  \text{subject to} & & T =\sum \nolimits_{i=1}^m \eta x_i,\\
& & & \chi(\eta \mathbf{x},\tilde{\mathcal{F}}) \in \mathcal{P}_{T}.
\label{eqn:minsnap-3}
\end{aligned}
\end{equation}
The feasibility constraint is typically evaluated using differential flatness of the idealized quadrotor dynamics~\cite{mellinger2011minimum}.
Specifically, the flatness transform provides a straightforward manner to obtain the control inputs, \ie, the motor speeds, that correspond to a quadcopter trajectory.
The feasibility of the trajectory can then be determined based on the admissibility of its required control inputs.

%


\subsection{Multi-Agent Trajectories}\label{sec:multiagenttraj}
In addition to the individual dynamics and obstacle avoidance constraints, multi-agent trajectory planning requires collision avoidance constraints for each pair of agents.
The problem is challenging because these constraints change as the trajectories of other vehicles are updated.

Trajectory discretization is a popular approach for formulating collision avoidance constraints.
By adding minimum-distance constraints between discrete points on each trajectory, the separation between the vehicles can be guaranteed.
Since the constraints are in quadratic form, nonlinear optimization is required to solve the problem, \eg,
by using a potential field method and iterative updates~\cite{ladinig2021time},
by utilizing sensitivity-based relaxation~\cite{wang2019game}, or
by formulating the mixed-integer linear programming (MILP) that approximates the quadratic constraints with multiple linear constraints and integer variables~\cite{mellinger2012mixed}.
Alternatively, collisions can be avoided by utilizing a leader-follower formation~\cite{oh2015survey}.
In this approach, a leader trajectory is first formulated, after which the remaining agents are set to follow the leader agent while keeping some formation.
Since only a single trajectory optimization is required, this method can relieve the stark increase in complexity, known as the \textit{curse of dimensionality}, caused by the joint optimization of multiple trajectories.
The leader-follower approach has been applied to various mobile robotics systems, such as 2D wheeled-robots~\cite{miao2018distributed}, helicopters~\cite{yun2008leader}, and quadrotors~\cite{ghamry2015formation}.
Recently, it was combined with deep reinforcement learning to generate a deep-neural-network-based leader-follower navigation system~\cite{deka2021hiding}.

In this paper, we compare our trajectory optimization results to two baseline methods: (i) a heuristic formation control scheme, and (ii) the MILP formulation from \cite{mellinger2012mixed}.
In the formation control scheme, vehicle collisions are avoided by enforcing that the vehicles keep a specified formation at all times.
We generate the trajectory of the formation center as well as trajectories for two parameters, corresponding to the scale and the rotation of the formation.
We specify a formation waypoint $k$ in terms of its center location, the maximum distance bound $\tilde{b}^\text{form}_{k}$ from this center to each of the vehicles, and the formation rotation, \ie, its yaw angle, $\tilde{\psi}^\text{form}_{k}$.
The piecewise polynomial trajectory of the formation scale parameter $b^\text{form}(t)$ is then generated by solving the following optimization:
\begin{equation}\label{eqn:planning_formation_control_profile}
\underset{b^\text{form}}{\text{minimize}} \;\; \int_{0}^{\sum \nolimits_{i=1}^{m} x_i} \norm{\frac{d^4 b^\text{form}}{dt^4}}^2
\end{equation}
\vspace{-1.2em}
\begin{multline*}
\text{subject to} \hfill b^\text{form}(\sum_{i=1}^{e_{k}} x_i) = \tilde{b}^\text{form}_k,\;k=1,\dots,N_f+1,\\
\hfill b^\text{form}(t) \leq \max(\tilde{b}^\text{form}_{k}, \tilde{b}^\text{form}_{k+1}),\\
\hfill \forall t\in\left[\sum \nolimits_{{j}=1}^{e_k}x_{{j}}, \sum \nolimits_{{j}=1}^{e_{k+1}}x_{{j}}\right],\;k=1,\dots,N_f, \\
\hfill b^\text{form}(0) = \tilde{b}^\text{form}_0, \\
\hfill b^\text{form}(t) \leq \max(\tilde{b}^\text{form}_{0},\tilde{b}^\text{form}_{1}),\forall t\in\left[0, \sum \nolimits_{{j}=1}^{e_1}x_{{j}}\right],\hspace{-1em}
\end{multline*}
where $e_k$ is the index of the $k$-th formation waypoint as in \eqref{eq:wpconstraint}, and $\tilde{b}^\text{form}_{0}$ and $\tilde{b}^\text{form}_{N_f+1}$ refer to the maximum distance bounds at the start and end points.
Using the formation scale profile, we generate the trajectory $p^\text{form}(t)$ for the formation center using \eqref{eqn:min_snap_smoothness} with added separation on the polytope collision avoidance constraints, as follows:
\begin{equation}
\begin{gathered}\label{eqn:planning_formation_control}
A_j p\left(t\right) \leq b_j + b^\text{form}(t) E_j, \\
\forall t\in\left[\sum \nolimits_{{i}=1}^{j-1}x_{{i}}, \sum \nolimits_{{i}=1}^{j}x_{{i}}\right], \; j=1,\cdots,m,
\end{gathered}
\end{equation}
where $E_j \in \{0,1\}^{d_j}$ with its $i$-the element set to zero if the trajectory passes through the $i$-th face of polytope $j$ and its remaining elements set to unity.
Similar to \eqref{eqn:planning_formation_control_profile}, we connect the formation yaw waypoints $\tilde{\psi}^\text{form}_{k}$ with a smooth polynomial $\psi^\text{form}(t)$ obtained by minimizing its second-order derivatives.
The time allocation is generated using \eqref{eqn:minsnap-2} with a modified objective function that includes the constraints from \eqref{eqn:planning_formation_control}.
Based on $b^\text{form}(t)$, $\psi^\text{form}(t)$, and $p^\text{form}(t)$, we generate the trajectory for each of the vehicles, and collectively slow down all trajectories until the corresponding required motor speeds remain within the feasible set, similar to \eqref{eqn:minsnap-3}.

In the MILP-based method, the collision constraints are included in the minimum-snap optimization as mixed-integer constraints.
The approach from \cite{mellinger2012mixed} formulates a MILP with the following vehicle collision avoidance constraints
\begin{equation}
\begin{aligned}\label{eqn:planning_milp}
|p_{i,k}(t) - p_{j,k}(t)| \geq d_\text{min} - M y_{i,j,k}, \; k\in\{x,y,z\}, \\
y_{i,j,k}\in\{0,1\}, \; i,j ={1,\dots,N_v},\;j > i, \\
\sum_{i,j,k} y_{i,j,k} \leq 3 N_v (N_v-1) / 2 - 1,
\end{aligned}
\end{equation}
where $M$ is a large number and $d_\text{min}$ is the minimum distance between vehicles.
In order to formulate the optimization problem as a linear program, $d_\text{min}$ is included as a component-wise constraint.
If two vehicles are separated diagonally, collision avoidance can be achieved with a component-wise smaller distance than $d_\text{min}$.
Therefore, in our implementation, we run a grid search on $d_\text{min}$ and adjust it separately for different obstacle configurations.

\subsection{Bayesian Optimization}
Bayesian optimization (BayesOpt) is a class of algorithms that can be used to solve optimization problems with unknown objective or constraint functions that are expensive to evaluate.
Evaluation points are selected to model the unknown functions and approach the optimum with maximum efficiency, so that the total number of evaluations is kept to a minimum.

Within the BayesOpt framework, Gaussian process classification (GPC) modeling is widely used to build a \textit{surrogate model} that approximates the unknown constraint functions.
Given data points $\mathbf{X}=\{\mathbf{x}_1,\cdots,\mathbf{x}_N\}$ with corresponding evaluations $\mathbf{y}=\{y_1,\cdots,y_N\}$, GPC assumes a joint Gaussian distribution of the evaluations and the latent variables $\mathbf{f}=\begin{bmatrix}f_1,\cdots,f_N\end{bmatrix}$, and predicts the probability $P(y_*|\mathbf{y},\mathbf{x}_*,\mathbf{X})$ for a test point $\mathbf{x}_*$ based on the latent variables.
These latent variables encode label probabilities for the evaluations, which can be obtained through a mapping onto the probability domain $[0,1]$, \eg, 
\begin{equation}
\Phi(x)=\int_{-\infty}^x \mathcal{N}(s|0,1)ds.
\end{equation}
The latent variables and the hyperparameters of the kernel function are trained by maximizing the marginal likelihood function
\begin{equation}
\begin{aligned}
P(\mathbf{y},\mathbf{f}|\mathbf{X})&=\Pi_{i=n}^N P(y_n|f_n)P(\mathbf{f}|\mathbf{X})\\&=\Pi_{n=1}^N \mathcal{B}(y_n|\Phi(f_n))\mathcal{N}(\mathbf{f}|0,K(\mathbf{X},\mathbf{X})),
\end{aligned}
\label{eqn:gpc_likelihood}
\end{equation}
where $\mathcal{B}(x)$ is the Bernoulli likelihood used to formulate $\Phi(f_n)$ as a probability distribution.
The covariance between $\mathbf{X}$ and a test point $\mathbf{x}_*$ is modeled with the same covariance kernel, as follows
\begin{gather}
    P(\mathbf{f},f_*|\mathbf{x}_*,\mathbf{X}) = \mathcal{N}\left(\begin{bmatrix}\mathbf{f}\\f_*\end{bmatrix}\left|0,
    \begingroup
    \setlength\arraycolsep{0.2pt}
    \begin{bmatrix}
        K(\mathbf{X},\mathbf{X}) & K(\mathbf{X},\mathbf{x}_*) \\
        K(\mathbf{x}_*,\mathbf{X}) & K(\mathbf{x}_*,\mathbf{x}_*)
    \end{bmatrix}
    \endgroup
    \right.\right),
    \label{eqn:gpc_1}
    \end{gather}
    so that the distribution of the latent variable $f_*$ can be estimated as
    \begin{gather}
    P(f_*|\mathbf{x}_*,\mathbf{X},\mathbf{y}) = \int P(f_*|\mathbf{f},\mathbf{x}_*,\mathbf{X})P(\mathbf{f}|\mathbf{X},\mathbf{y})d\mathbf{f}. \label{eqn:gpc_2}
\end{gather}
The resulting class probability is obtained by
\begin{gather}
P(y_*|\mathbf{x}_*,\mathbf{X},\mathbf{y}) = \int P(y_*|f_*)P(f_*|\mathbf{x}_*,\mathbf{X},\mathbf{y})d\mathbf{f_*}.
\label{eqn:gpc_3}
\end{gather}
For more details on GPC and its implementation, the reader is referred to \cite{nickisch2008approximations}.

The BayesOpt \textit{acquisition function} is designed to select the next evaluation point by considering both reducing the uncertainty of the surrogate model and finding the precise optimum of the objective function.
Based on the data $\mathcal{D}$ obtained in previous evaluations, the surrogate model can be trained to approximate the unknown function.
Next, the approximate optimal solution can be estimated using the trained surrogate model.
Each next evaluation point is obtained by solving the following optimization problem:
\begin{equation}
    \mathbf{x}_\text{next} = \argmax_{\mathbf{x}} \alpha(\mathbf{x}|\mathcal{D})
    \label{eqn:bg_acquisition_func}
\end{equation}
where $\alpha(\mathbf{x}|\mathcal{D})$ is the acquisition function that represents the value of an evaluation point $\mathbf{x}$ given the data $\mathcal{D}$.
If the surrogate model is used to approximate an unknown constraint function, the acquisition function can be based on, \eg, the product of the expected objective improvement and the probability of constraint satisfaction \cite{gardner2014bayesian, gelbart2015constrained}, or the expected entropy reduction of the distribution over the solution domain that satisfies the constraints \cite{hernandez2015predictive, marco2019classified}.
In our proposed algorithm, Bayesian optimization is applied to model the dynamic feasibility and collision avoidance constraints of the multi-agent planning problem.

Multi-fidelity Bayesian optimization combines function evaluations of different fidelity levels.
The key idea is that combining cheap low-fidelity evaluations with expensive high-fidelity measurements improves overall efficiency.
To incorporate information from multiple sources, the surrogate model must be modified to combine multi-fidelity evaluations, \eg, using a linear transformation to describe the relationship between different fidelity levels~\cite{kennedy2000predicting, le2014recursive}. 
Suppose that we have $L$ fidelity levels, and each level is denoted by $l \in \{l^1,l^2,\dots,l^L\}$, where $l^1$ is the level of the lowest-fidelity experiment and $l^L$ is the level of the highest-fidelity experiment,
then the relationship between adjacent fidelity levels $l^j$ and $l^{j-1}$ can be modeled as
\begin{equation}
    f_{l^j}(\mathbf{x}) = \rho_{l^{j-1}} f_{l^{j-1}} (\mathbf{x}) + \delta_{l^j} (\mathbf{x}),
    \label{eqn:bg_multi_fidelity_linear}
\end{equation}
where $f_{l^j}(\mathbf{x})$ and $f_{l^{j-1}} (\mathbf{x})$ are the output distributions of $\mathbf{x}$ for $l^j$ and for $l^{j-1}$, respectively.
The bias distribution $\delta_{l^j}$ is independent of $f_{l^{j-1}}, \dots, f_{l^1}$, and the constant $\rho_{l^{j-1}}$ represents the correlation between the output distributions for the two adjacent fidelity levels.
Similar to the surrogate model, the acquisition function has to be modified to incorporate multi-fidelity evaluations.
In the multi-fidelity Bayesian optimization framework, the acquisition function is used not only to select the next evaluation point, but also its fidelity level, as follows:
\begin{equation}
    \mathbf{x}_\text{next}, l_\text{next} = \argmax_{\mathbf{x}, l\in\{l^1,\dots,l^L\}} \alpha(\mathbf{x},l|\mathcal{D}).
    \label{eqn:bg_multi_fidelity_acquisition_func}
\end{equation}
The acquisition function itself is modified by introducing weights based on the evaluation cost at the different fidelity levels.
In practice, high-fidelity evaluations will have smaller weights compared to low-fidelity evaluations.
This makes the algorithm less likely to select high-fidelity evaluations, so that the overall cost of the experiments is minimized.


In practice, the acquisition function is often discontinous or nonlinear and cannot be solved analytically.
BayesOpt generates and evaluates (random) candidate solutions to select the acquisition function optimum.
The required number of candidate solutions increases exponentially with problem dimension.
Consequently, BayesOpt for high-dimensional problems is computationally expensive and often infeasible due to the prohibitive cost of solving the acquisition function.
In multi-agent planning, the dimension of the solution space increases linearly with the number of vehicles, leading to an exponential increase in candidate solutions.
Recent works on high-dimensional BayesOpt focus on decomposing the high-dimensional space and solving the acquisition function on each subspace separately.
For example,
\cite{kandasamy2015high, rolland2018high} assume the acquisition function has additive structure so that the state space can be linearly decomposed into low-dimensional subspaces, and
\cite{li2017high, kirschner2019adaptive} iteratively select active subsets of variables and optimize the acquisition function only on the selected low-dimensional subspace.
In our proposed algorithm, we employ a similar approximation on the high-dimensional solution space, where the decomposed acquisition functions correspond to the dynamic feasibility constraints of individual trajectories (first-order) and to the pairwise collision avoidance constraints of pairs of trajectories (second-order).


\section{Algorithm} \label{sections:algorithm}

We propose a BayesOpt framework to efficiently minimize the total trajectory time $\max_{i=1} T_i$ of a cooperative multi-agent trajectory using multi-fidelity evaluations.
We leverage \eqref{eqn:min_snap_smoothness} to obtain a mapping from the time allocation over segments $\mathbf{x}_i$ to a smooth minimum-snap trajectory $p_i = \chi(\mathbf{x}_i, \tilde{\mathcal{F}_i})$ that satisfies the obstacle avoidance and formation waypoint constraints.
This mapping enables us to transform the infinite-dimensional trajectory generation problem to the problem of finding the optimal multi-agent time allocation over segments.
Based on this approach, we reformulate the general multi-agent minimum-time planning problem \eqref{eqn:planning_general}, as follows:
\begin{equation}\label{eqn:alg_problem_formulation}
\underset{\mathbf{x} \in\realR^{m \times N_v}_{\geq 0}, \mathbf{T}}{\text{minimize}} \max_{i=1,\dots,N_v} T_i
\end{equation}
\vspace{-2em}
\begin{multline*}
\text{subject to} \hfill\\
T_i = \sum \nolimits_{j=1}^m x_{ij}, \hfill i=1,\dots,N_v, \\
\sum_{j=1}^{e_{k}} x_{i,j} = \sum_{j=1}^{e_{k}} x_{\tilde i,j}, \hfill i,\tilde i=1,\dots,N_v,\; k =1,\dots,N_f\\
\chi(\mathbf{x}_i, \tilde{\mathcal{F}_i}) \in \mathcal{P}_{T_i}, \hfill i=1,\dots,N_v, \\
\left(\chi(\mathbf{x}_i, \tilde{\mathcal{F}_i}), \chi(\mathbf{x}_j, \tilde{\mathcal{F}_j})\right) \in \mathcal{F}_{T_i,T_j}, \hfill i,j =1,\dots,N_v,\;j > i.
\hspace{-1em}
\end{multline*}
The first constraint in \eqref{eqn:alg_problem_formulation} is linear and defines the total trajectory time.
The second constraint, also linear, enforces that all formation waypoints are attained synchronously.
The two remaining constraints enforce dynamic feasibility of each trajectory and collision avoidance between each pair of trajectories, respectively.
As described in Section \ref{sections:introduction}, it is challenging to verify these two final constraints.
Dynamic feasibility requires that a trajectory can be flown on the actual vehicles and depends on the complex dynamics of fast and agile flight in tight formations.
Existing work has already demonstrated that Bayesian optimization can be used to efficiently model this dynamic feasibility constraint~\cite{ryou2021ijrr}.

Our major contribution pertains to the handling of the collision avoidance constraint. 
Since this constraint couples the optimization variables from different vehicles, the trajectory optimization should be solved jointly, resulting in a high-dimensional input domain.
As the volume of the data space increases exponentially with dimension, the required amount of data also increases rapidly.
This curse of dimensionality prevents the application of standard methods for BayesOpt.
In this section, we provide an overview of our proposed modular Bayesian optimization algorithm with particular focus on how we address high dimensionality in the definition of the surrogate model and the acquisition function.

\begin{algorithm}
\DontPrintSemicolon
\LinesNumbered
\SetSideCommentRight
\KwIn{\begin{tabular}[t]{l}
        Surrogate model $\mathcal{M}_1, \cdots, \mathcal{M}_{N_v}$, \\
        $\mathcal{M}_{1,2}, \cdots, \mathcal{M}_{N_v-1,N_v}$, \\
        acquisition function $\alpha$, \\
        size of candidate data points $N_s$
    \end{tabular}
}
\SetKwProg{sFn}{subfunction}{:}{}
\SetKwFunction{KwsFn}{SampleTraj}
\sFn{\KwsFn{$i$, $\mathcal{X}_F$}}{
    $\mathcal{X} = \emptyset$\;
    \While{$|\mathcal{X}| < N_1$}{
        $\mathcal{X}_t \gets \text{Randomly sampled }N_s\text{ points}$ \;
        Rescale $\mathcal{X}_t$ with $\mathcal{X}_F$ \;
        Remove $\mathbf{x}\in\mathcal{X}_t$ s.t. $\Tilde{P}_i(y_i=1|\mathbf{x},\mathcal{M}_i) < C_1$\;
        $\mathcal{X}\gets\mathcal{X}\cup\mathcal{X}_t$\;
    }
    \KwRet $\mathcal{X}$\;
}
\SetKwProg{Fn}{function}{:}{}
\SetKwFunction{KwFn}{Modular Bayes-Opt}
\Fn{\KwFn{}}{
\Repeat{convergence of solution}{
    $\mathcal{X}_F \gets \text{Randomly sampled }N_s\text{ points}$\;
    $\mathcal{X} = \mathcal{X}_F$\;
    \While{$|\mathcal{X}| < N_2$}{
        $\mathcal{X}_t \gets$ SampleTraj(1, $\mathcal{X}_F$)\;
        \For{$i=2,\cdots,N_v$}{
            $\mathcal{X}_t\gets\mathcal{X}_t\times$SampleTraj($i$, $\mathcal{X}_F$)\;
            \For{$j=1,\cdots,i-1$}{
                Remove $\mathbf{x}\in\mathcal{X}_t$ s.t. $\Tilde{P}_{j,i}(y_{j,i}=1|\mathbf{x},\mathcal{M}_{j,i}) < C_2$\;
            }
        }
        $\mathcal{X}\gets\mathcal{X}\cup\mathcal{X}_t$\;
    }
    $\mathbf{x}, l \gets\argmax_{\mathbf{x}\in\mathcal{X}, l\in\{l^1,\cdots,l^L\}} \alpha(\mathbf{x}, l|\mathcal{D})$\;
    Evaluate $\mathbf{x}$ in $l$-th fidelity experiment\;
    Update dataset $\mathcal{D}_{l,i}, \mathcal{D}_{l,i,j}$\;
    Update the surrogate model $\mathcal{M}_{l,1},\cdots,\mathcal{M}_{l,N_v}, \mathcal{M}_{l,1,2},\cdots,\mathcal{M}_{l,N_v-1,N_v}$\;
}
$p_i^* \gets \chi(\mathbf{x}_i^*, \tilde{\mathcal{F}_i})\: \forall i=1,\cdots,N_v$\;
$T^* \gets \max_{i=1,\cdots,N_v} \sum \nolimits_{j=1} x^*_{ij}$\;
\KwOut{$T^*$, $\left[p_1^*,\cdots,p_{N_v}^*\right]$}
}
\caption{Modular Bayesian optimization with simultaneous acquisition function evaluation}
\label{alg:main}
\end{algorithm}

\subsection{Modular Surrogate Model}
In multi-agent planning, the number of variables increases proportionally to the number of vehicles.
Consequently, surrogate model training and evaluation time increase rapidly as more vehicles are considered.
Moreover, modeling performance may suffer, even when provisions are made to handle a large amount of data points, such as by using the inducing points technique as in \cite{hensman2015scalable}.
For example, in experiments, we noticed that a surrogate model may learn only part of the constraints, \eg, it may represent the dynamic feasibility constraints well but neglect the collision avoidance constraints.

In this work, we propose a modular surrogate model, shown in Fig. \ref{fig:diagram_main_algorithm}, to address these challenges. 
Our model is comprised of several GPC dynamic feasibility constraints and collision models.
Each surrogate model $\mathcal{M}$ consists of the latent variables $\mathbf{f}=\begin{bmatrix}f_1&\cdots,f_N\end{bmatrix}$ and the hyperparameters $\theta$ of the covariance matrix, such that $\mathcal{M}=\left(\mathbf{f},\theta\right)$.
%
We denote the dataset consisting of trajectories for vehicle $i$ by $\mathcal{D}_i$.
The corresponding surrogate model $\mathcal{M}_i$, trained with $\mathcal{D}_i$, approximates the constraints that apply solely to vehicle $i$, \ie, the obstacle avoidance constraint $\mathcal{F}_{T_i}$ and the dynamic feasibility constraint $\mathcal{P}_{T_i}$ in \eqref{eqn:planning_general}.
Similarly, the dataset $\mathcal{D}_{i,j}$ contains multi-agent trajectories for vehicles $i$ and $j$, and the corresponding surrogate model $\mathcal{M}_{i,j}$, trained with $\mathcal{D}_{i,j}$, approximates the collision avoidance constraint for vehicles $i$ and $j$, \ie, $\mathcal{F}_{T_i,T_j}$ in \eqref{eqn:planning_general}.

In order to leverage evaluations at $L$ fidelity levels, we further expand the definition of the surrogate model by defining the model at fidelity level $l$ as
\begin{align}
    \mathcal{M}_{l} = \left\{\mathcal{M}_{l, 1},\dots,\mathcal{M}_{l, N_v},\mathcal{M}_{l, 1, 2},\dots,\mathcal{M}_{l, N_v-1, N_v}\right\}.
    \label{eqn:alg_mfbo_surrogate}
\end{align}
We then use the multi-fidelity deep Gaussian process (MFDGP) from \cite{cutajar2019deep} as covariance kernel function to estimate the multi-fidelity GP prior.
Uncertainty quantification computations are accelerated using the inducing points method.

\subsection{Acquisition Function}
We use the acquisition function from \cite{ryou2021ijrr} to consider both \textit{exploration} to improve the surrogate model and \textit{exploitation} to find the optimal solution.
In exploration, we select the most uncertain sample near the decision boundary of the classifier~\cite{costabal2019multi}.
Since the latent function mean approaches zero at the decision boundary, this sample is found as the maximizer of
\begin{align}
    \alpha_\text{explore}(\mathbf{x}, l) = -\sum_{i=1}^{N_v} \frac{|\mu_l(\mathbf{x}_{i})|}{\sigma_l(\mathbf{x}_{i})} - \sum_{i=1}^{N_v-1}\sum_{j=i+1}^{N_v} \frac{|\mu_l(\mathbf{x}_{i},\mathbf{x}_{j})|}{\sigma_l(\mathbf{x}_{i},\mathbf{x}_{j})} ,
    \label{eqn:alg_acquisition_explore}
\end{align}
where $\left(\mu_l(\mathbf{x}_{i}), \sigma_l(\mathbf{x}_{i})\right)$ and $\left(\mu_l(\mathbf{x}_{i}, \mathbf{x}_{j}), \sigma_l(\mathbf{x}_{i}, \mathbf{x}_{j})\right)$ are the mean and standard deviation of the $l$-th fidelity models' posterior distributions $P(f|\mathbf{x}_{i}, \mathcal{D}_{l,i})$ and $P(f|\mathbf{x}_{i}, \mathbf{x}_{j}, \mathcal{D}_{l, i,j})$ obtained from \eqref{eqn:gpc_2}.

In exploitation, we utilize expected improvement with constraints (EIC) to quantify the expected effectiveness of a candidate data point based on the product of expected objective improvement and the probability of feasibility \cite{gardner2014bayesian}.
In order to discourage overly optimistic evaluations, we modify EIC to not only consider the probability of success, but also the corresponding variance, as follows:
\begin{align}
    &\Tilde{P}(y_{l,i}=1|\mathbf{x}_{i}) = P(\mu_l(\mathbf{x}_{i}) - \beta_l\sigma_l(\mathbf{x}_{i}) \geq 0|\mathbf{x}_{i}), \\
    &\Tilde{P}(y_{l,i,j}=1|\mathbf{x}_{i},\mathbf{x}_{j}) = \nonumber \\
    &\:\:\:\:\:\:\:\:\: P(\mu_l(\mathbf{x}_{i},\mathbf{x}_{j}) - \beta_l\sigma_l(\mathbf{x}_{i},\mathbf{x}_{j}) \geq 0|\mathbf{x}_{i},\mathbf{x}_{j}),
    \label{eqn:alg_acquisition_exploit_lower_bound}
\end{align}
where $\beta_l$ is the penalty weight on the variance.
The probability that a set $\mathbf{x}$ of time allocations for all vehicles is feasible is computed as 
\begin{equation}
\Tilde{P}_l(y=1|\mathbf{x}) = \left(\prod_{i=1}^{N_v} \Tilde{P}_{l,i}\right) \:\:\left(\prod_{i=1}^{N_v-1}\prod_{j=i+1}^{N_v} \Tilde{P}_{l,i,j}\right),
\end{equation}
where $\Tilde{P}_{l,i} = \Tilde{P}(y_{l,i}=1|\mathbf{x}_{i})$ and $\Tilde{P}_{l,i,j}=\Tilde{P}(y_{l,i,j}=1|\mathbf{x}_{i}, \mathbf{x}_{j})$.
The resulting acquisition function is then given by
\begin{equation}
    \alpha_\text{exploit}(\mathbf{x},l) = 
    \begin{cases}
        \alpha_{EI}(\mathbf{x})\Tilde{P}_l(y=1|\mathbf{x}), &\text{if } \forall i\neq j, \Tilde{P}_{l,i} \geq h_l, \\
                                                            &\:\:\:\:\Tilde{P}_{l,i,j} \geq h_l \\
        0, &\text{otherwise}
    \end{cases}
    \label{eqn:alg_acquisition_exploit}
\end{equation}
where the bound $h_l$ serves to avoid evaluations that are likely infeasible.
Since the objective function is deterministic, so is the expected improvement, given by
\begin{equation}
\alpha_{EI}(\mathbf{x}) = \max_{i=1,\cdots,N_v} \left(\sum_j \Bar{x}_{ij}\right) - \max_{i=1,\cdots,N_v} \left(\sum_j x_{ij} \right)
\end{equation}
where $x_{ij}$ is a $j$-th element of $i$-th vehicle's time allocation $\mathbf{x}_{i}$, and $\bar{\mathbf{x}}$ is the current best solution.

Finally, we combine \eqref{eqn:alg_acquisition_explore} and \eqref{eqn:alg_acquisition_exploit} to obtain
\begin{equation}
    \alpha(\mathbf{x}, l) =  
    \begin{cases}
        \alpha_\text{exploit}(\mathbf{x}, l), &\text{if } \exists \textbf{x}\in\mathcal{X} \:\text{s.t.}\:\alpha_\text{exploit}(\textbf{x}, l) > 0 \\
        \alpha_\text{explore}(\mathbf{x}, l), &\text{otherwise}
    \end{cases}
    \label{eqn:alg_acquisition}
\end{equation}
where $\mathcal{X}$ is the set of sample trajectories generated in Algorithm \ref{alg:main}.

\subsection{Simultaneous Acquisition Function Evaluation and Random Sampling}
It is often not possible to find the next data point and its fidelity level by explicitly solving \eqref{eqn:bg_multi_fidelity_acquisition_func}.
Instead, a typical course in BayesOpt is to first sample a set of candidate data points and then evaluate the acquisition function at all of these data points in order to select the next evaluation point.
This method works well in many practical scenarios, but may be problematic when optimizing over a high-dimensional space because the amount of candidate data points required to obtain a reliable result increases exponentially with the dimension.
Moreover, the computational burden of each evaluation of the acquisition function, which requires evaluation of \eqref{eqn:gpc_3}, increases with the size of the surrogate model and may even surpass the cost of actual evaluation of the data point with regard to the objective and constraints of \eqref{eqn:alg_problem_formulation}.

In order to improve sampling efficiency, we propose to combine the random sampling and acquisition function evaluation steps.
This procedure and the resulting BayesOpt framework are detailed in Algorithm \ref{alg:main}.
First, we generate a set $\mathcal{X}_F$ of random samples for the time allocation between the formation waypoints.
Then, we sample the candidate data points for each vehicle sequentially and obtain the posterior distribution $P(y_i=1|\mathbf{x}_i, \mathcal{D}_i)$, which is used in the acquisition function evaluation.
When sampling the candidate data points, $\mathcal{X}_F$ is used to rescale the time allocations for each vehicle in order to ensure that the formation waypoints are attained synchronously.
Next, the data points with low posterior probability, \ie, those that are least likely to satisfy the dynamic feasibility constraint, are removed from the set of candidate points.
Using the remaining data points, we generate pairwise combinations and estimate the distribution $P(y_{i,j}=1|\mathbf{x}_i, \mathbf{x}_j, \mathcal{D}_{i,j})$ corresponding to the probability that the collision avoidance constraints are satisfied.
Again, we remove data points with low posterior probability.
The parameters $C_1$ and $C_2$ correspond to the cutoff levels of the rejection sampling steps and are adjusted based on the sample acceptance rate of the previous iteration.

Once the set $\mathcal{X}$ of candidate data points is obtained, the next evaluation points are selected based on \eqref{eqn:alg_acquisition}.
At each iteration, the number of evaluation points is adjusted based on the computational cost of the evaluations in the previous iteration.
The evaluated points are decomposed into and appended to the datasets $\mathcal{D}_i$ and $\mathcal{D}_{i,j}$ that are used to update the surrogate models $\mathcal{M}_i$, $\mathcal{M}_{i,j}$.
We use the minimum-jerk approach proposed in \cite{ryou2021ijrr} to generate smooth samples at all random sampling steps.


\subsection{Initialization and Pre-training}
An initial best solution is found by solving \eqref{eqn:minsnap-2} and \eqref{eqn:minsnap-3} for each vehicle individually.
Since the resulting solution does not satisfy the constraint that all vehicles attain the formation waypoints at the same time, we adjust the time allocation for segment $j$ along the trajectory for vehicle $i$ as follows:
\begin{equation}
x_{i,j} \leftarrow x_{i,j} \frac{N_v\sum_{\tilde{j}=e_{k}}^{e_{k+1}} x_{i,\tilde{j}}}{\sum_{i=1}^{N_v} \sum_{\tilde{j}=e_{k}}^{e_{k+1}} x_{i,\tilde{j}}},
\end{equation}
where $j \in [e_k,e_{k+1})$, so that all vehicles simultaneously attain the next formation waypoint.
Finally, we uniformly increase the time allocation to slow down the trajectories until all satisfy the dynamic feasibility constraints and obstacle avoidance constraints again.
We note that the resulting initial solution is not necessarily feasible, because it may not satisfy the collision avoidance constraints.
However, we found that it does represent a good starting data point around which we sample to obtain the initial data set and build the surrogate models.

We also use the initial best solution for the normalization of the data points.
By scaling the time allocation for each trajectory segment using the corresponding value in the initial solution, we maintain solution values of a similar scale, which improves the numerical stability of the training process.

\section{Experimental Results} \label{sections:experiment}

\begin{figure*}
	\centering
    \includegraphics[width=\textwidth,trim=.0cm .0cm .0cm .0cm,clip]{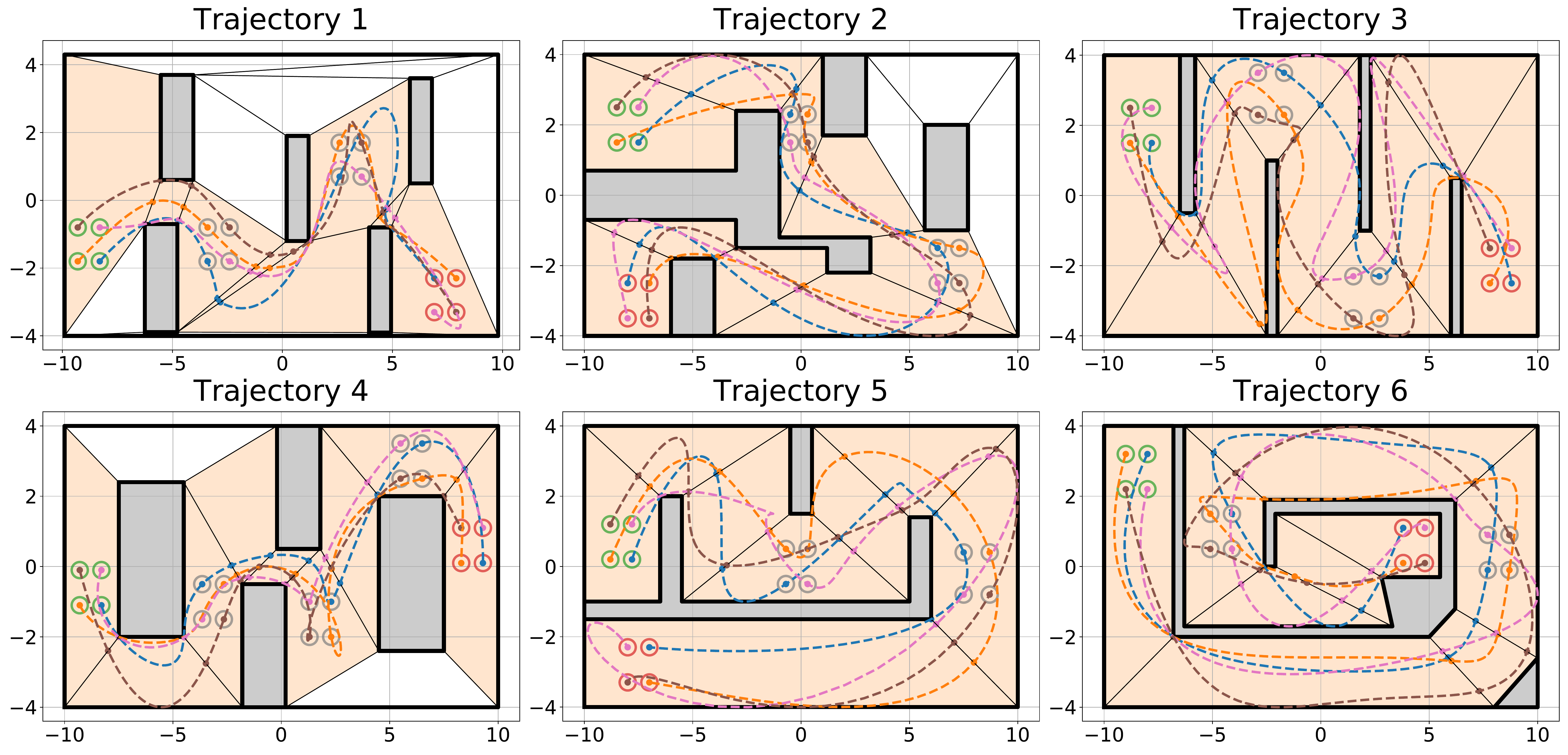}
    \captionsetup{width=0.9\textwidth, justification=centering}
	\caption{Single-fidelity trajectory optimization results. Start and end points are indicated by green and red circles, respectively. The synchronized formation waypoints are indicated by gray circles.}
	\label{fig:exp_trajectories}
\end{figure*}

The proposed algorithm is evaluated in various computational and real-world experiments.
First, we focus on generating multi-agent trajectories using single-fidelity Bayesian optimization.
We use a simple feasibility check based on the flatness transform, as described in Section \ref{sec:quadplan}.
The resulting trajectories, which include challenging obstacle configurations, are compared to results from the two baseline methods described in Section \ref{sec:multiagenttraj}.

Next, we extend the experiments to multi-fidelity Bayesian optimization.
Evaluations based on differential flatness now serve as a low-fidelity dataset, while a more realistic 6-DOF flight dynamics simulation is used to generate high-fidelity data.
The results demonstrate that our algorithm is capable of learning the direction of the anticipated trajectory-tracking error and considers this direction when modeling the collision avoidance constraints.

In both the single-fidelity and the multi-fidelity experiments, we use the six environments with obstacles shown in Fig. \ref{fig:exp_trajectories} and obtain dynamically feasible multi-agent trajectories that synchronously attain the specified formation waypoints and avoid obstacles and collisions.
We validated all six trajectories in real-world flight experiments using four quadcopter vehicles in a motion capture space.
A video with animations and recordings of the computational and real-world experiments is included as supplementary material.


\subsection{Single-Fidelity Optimization}
\begin{figure}
	\centering
	\includegraphics[width=0.48\textwidth,trim=.0cm .0cm .0cm .0cm,clip]{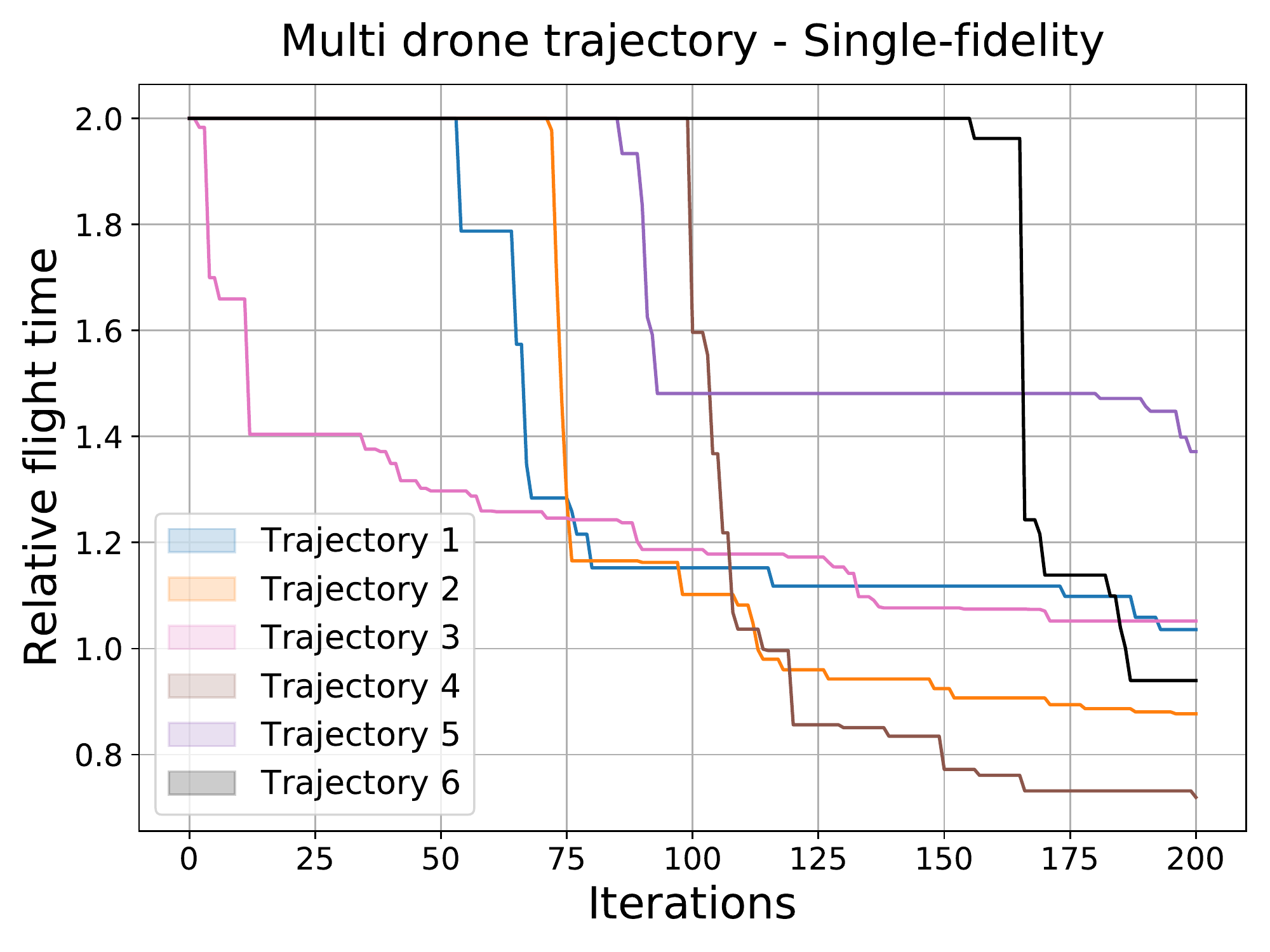}
	\caption{Relative flight time of single-fidelity optimized trajectories compared to the initial best solution.}
	\label{fig:exp_sta_opt}
\end{figure}

In the single-fidelity optimization we verify satisfaction of the collision avoidance constraints based on the planned trajectories and of the dynamic feasibility constraints based on the motor speeds obtained using the differential flatness transform.
We set the parameters of the acquisition functions \eqref{eqn:alg_acquisition_exploit_lower_bound} and \eqref{eqn:alg_acquisition_exploit} to $C_{l_1} = 1$, $h_{l_1} = 0.001$ and $\beta_{l_1} = 3.0$ and initialize the adaptive sample rejection thresholds $C_1$ and $C_2$ to 0.8.
Tuning these parameters mainly affects the computation time and not the actual optimization result, since the parameters mainly affect the efficiency the sampling process. 
Besides the extreme case where every random sample is rejected, the accepted samples update the surrogate model to arrive at the same optimal solution.
At each iteration, these thresholds are adjusted by 1\% until an acceptance rate of 0.25 is achieved.
The minimum distance between vehicles is set to 40 \si{cm}, measured from centroid to centroid.
For each of the six environments, we run 200 iterations of Bayesian optimization.
At each iteration, the 128 samples with the highest expected improvements based on \eqref{eqn:alg_acquisition} are selected for evaluation.
The relative flight time with regard to the initial best solution (which does not consider collision avoidance) for each iteration is shown in Fig. \ref{fig:exp_sta_opt}, and the final optimized trajectories are shown in Fig. \ref{fig:exp_trajectories}.

The optimized multi-agent trajectory is defined by the time allocation over the trajectory segments for each vehicle.
Fig. \ref{fig:exp_trajectories} shows that each trajectory exploits the polytope shapes to avoid collision with other vehicles and obstacles.
By increasing the time allocated to a segment, the trajectory can be shifted or a loop can be introduced (\eg, in Trajectory 4) to allow other vehicles to pass.
Similarly, in the later part of Trajectory 3, two of the agents (the brown and pink trajectories) perform large turns, which allows the other two agents to ``catch up'' so that all agents reach the final formation waypoint at the same time. 
The wide turns allow the vehicles to ``wait'' with minimum extra control effort.
In addition to the time allocation, we also attempted to optimize the polytope entry and exit points.
However, this requires three times more optimization variables and was found to result in inferior results for the same number of iterations.

\begin{figure}
	\centering
	\begin{subfigure}[b]{0.235\textwidth}
		\includegraphics[width=\textwidth,trim=4.0cm 3.0cm 4.cm 3.0cm,clip]{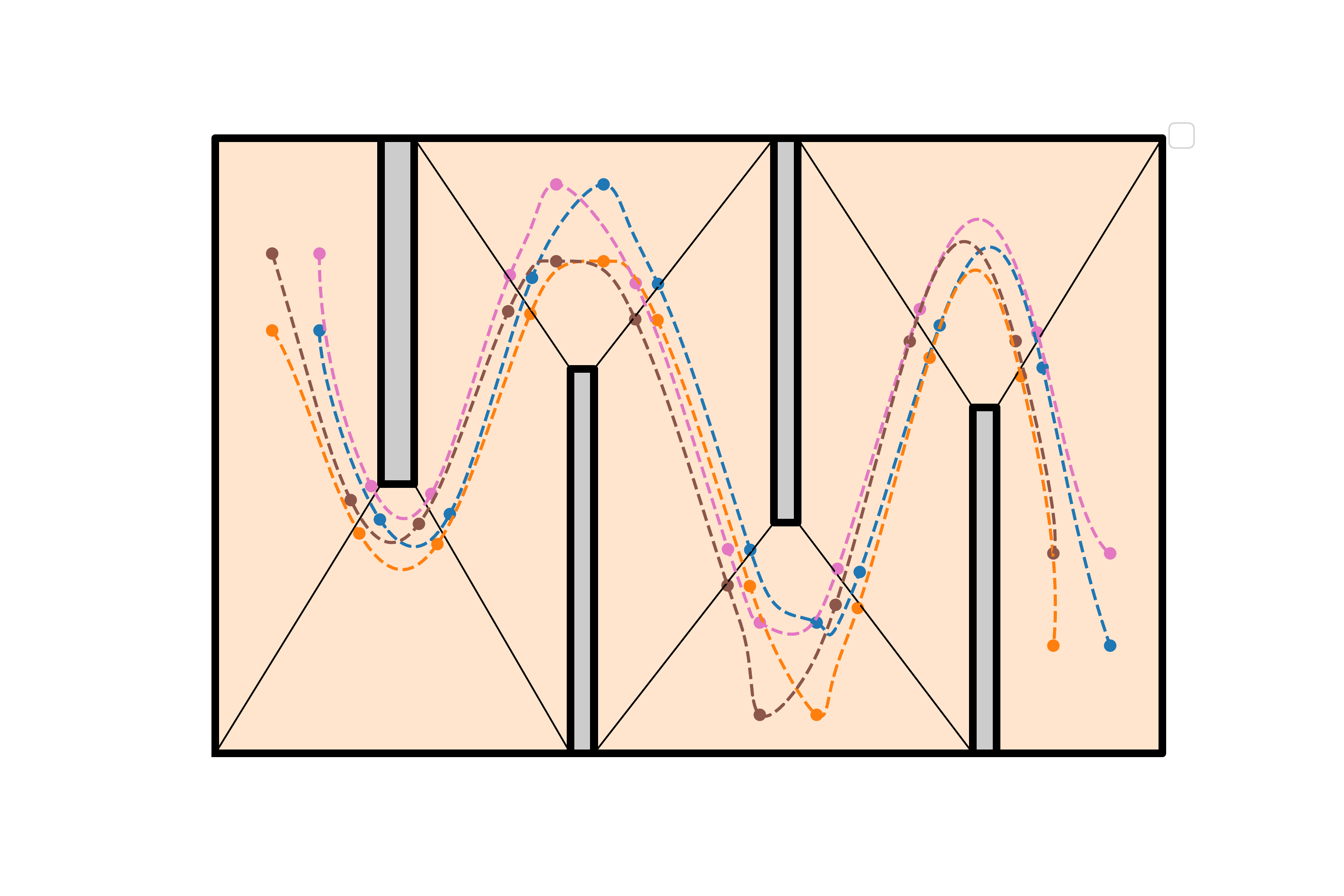}
		\caption{Formation control.}
		\label{fig:exp_formation_control}
	\end{subfigure}
	\begin{subfigure}[b]{0.235\textwidth}
		\includegraphics[width=\textwidth,trim=4.0cm 3.0cm 4.cm 3.0cm,clip]{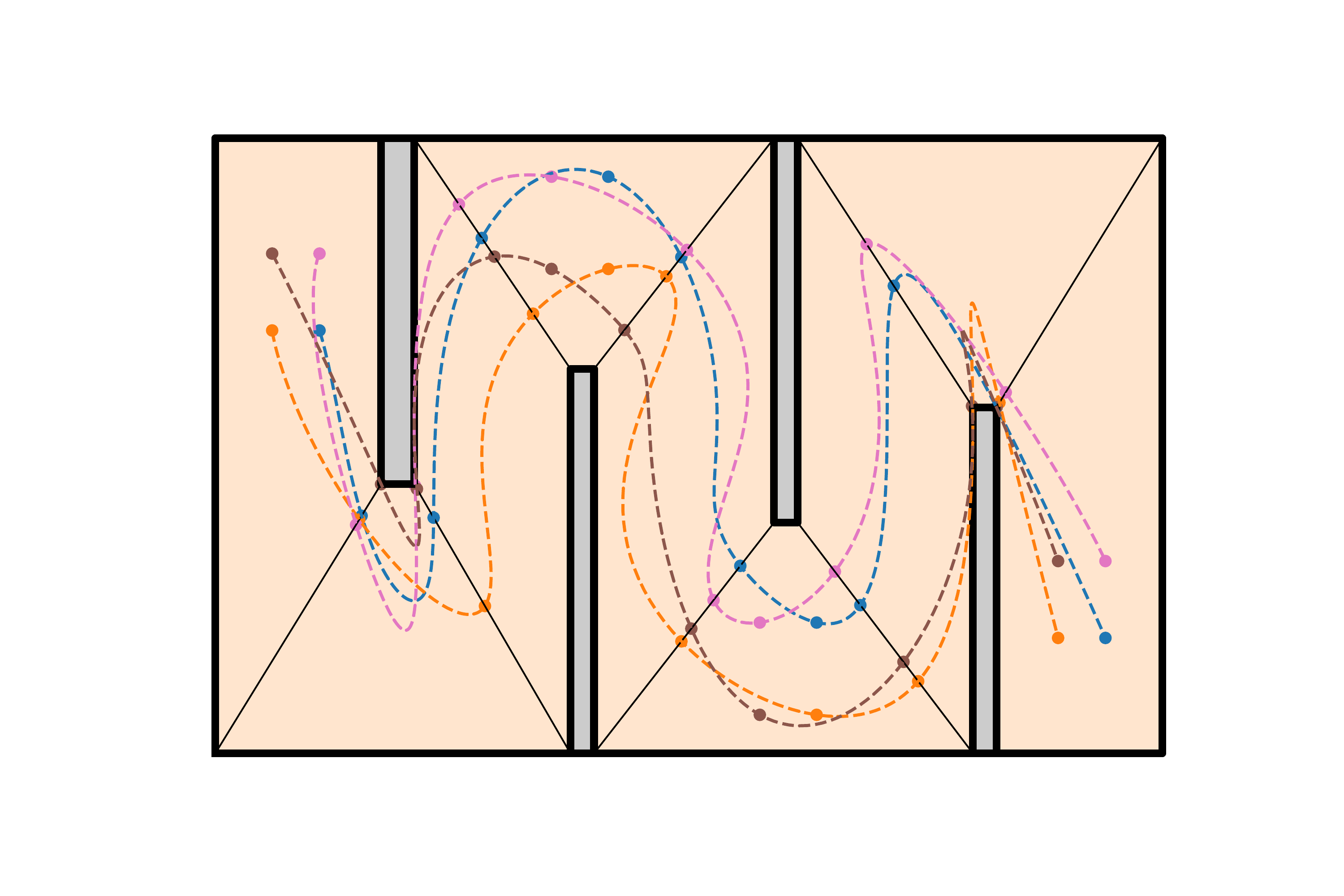}
		\captionsetup{width=0.9\textwidth, justification=centering}
		\caption{MILP.}
		\label{fig:exp_milp}
	\end{subfigure}
	\caption{Baseline results for Trajectory 3.}
	\label{fig:exp_comparsion}
\end{figure}

We compare the optimization results to the two baseline methods described in Section \ref{sec:multiagenttraj}: formation control and the MILP-based algorithm from \cite{mellinger2012mixed}.
The formation control method reliably generates feasible trajectories, but the resulting trajectories slow down significantly while the formation is scaling or rotating.
In contrast, the MILP-based algorithm generates typically faster trajectories.
However, since the MILP requires a large number of optimization variables, the optimization algorithm often fails to find a solution.
Table \ref{tab:exp_alg_compare} presents the trajectory times obtained using the baseline methods and the proposed algorithm.
Figure \ref{fig:exp_comparsion} clearly shows how the baseline algorithms result in different trajectories.
When comparing Fig. \ref{fig:exp_milp} to the corresponding trajectory in Fig. \ref{fig:exp_trajectories}, we notice that our proposed algorithm is able to generate a faster trajectory because of its capability to incorporate less conservative collision avoidance constraints.

\begin{table}
\small\centering
\caption{Trajectory times obtained from formation control, MILP \cite{mellinger2012mixed}, and the proposed modular Bayesian optimization (mBO). Min-snap corresponds to the initial solution obtained from \eqref{eqn:minsnap-2} and \eqref{eqn:minsnap-3} without considering the collision avoidance constraint.}
\label{tab:exp_alg_compare}
\begin{tabular}{lllll}
\hline
& Min-snap & Formation & MILP & mBO (\textbf{Ours})  \\
\hline
Traj. 1 & 5.359 s & 14.767 s & Failed & 5.548 s \\
Traj. 2 & 6.484 s & 29.794 s & 12.346 s & 5.686 s \\
Traj. 3 & 7.693 s & 14.993 s & 9.285 s & 8.093 s \\
Traj. 4 & 7.256 s & 16.375 s & 6.951 s & 5.221 s \\
Traj. 5 & 5.877 s & 16.050 s & 9.608 s & 8.061 s \\
Traj. 6 & 8.206 s & 34.500 s & 11.083 s & 7.713 s \\
\hline
\end{tabular}
\end{table}

\subsection{Multi-Fidelity Optimization}
\begin{figure}
	\centering
	\includegraphics[width=0.46\textwidth,trim=.0cm .0cm .0cm .0cm,clip]{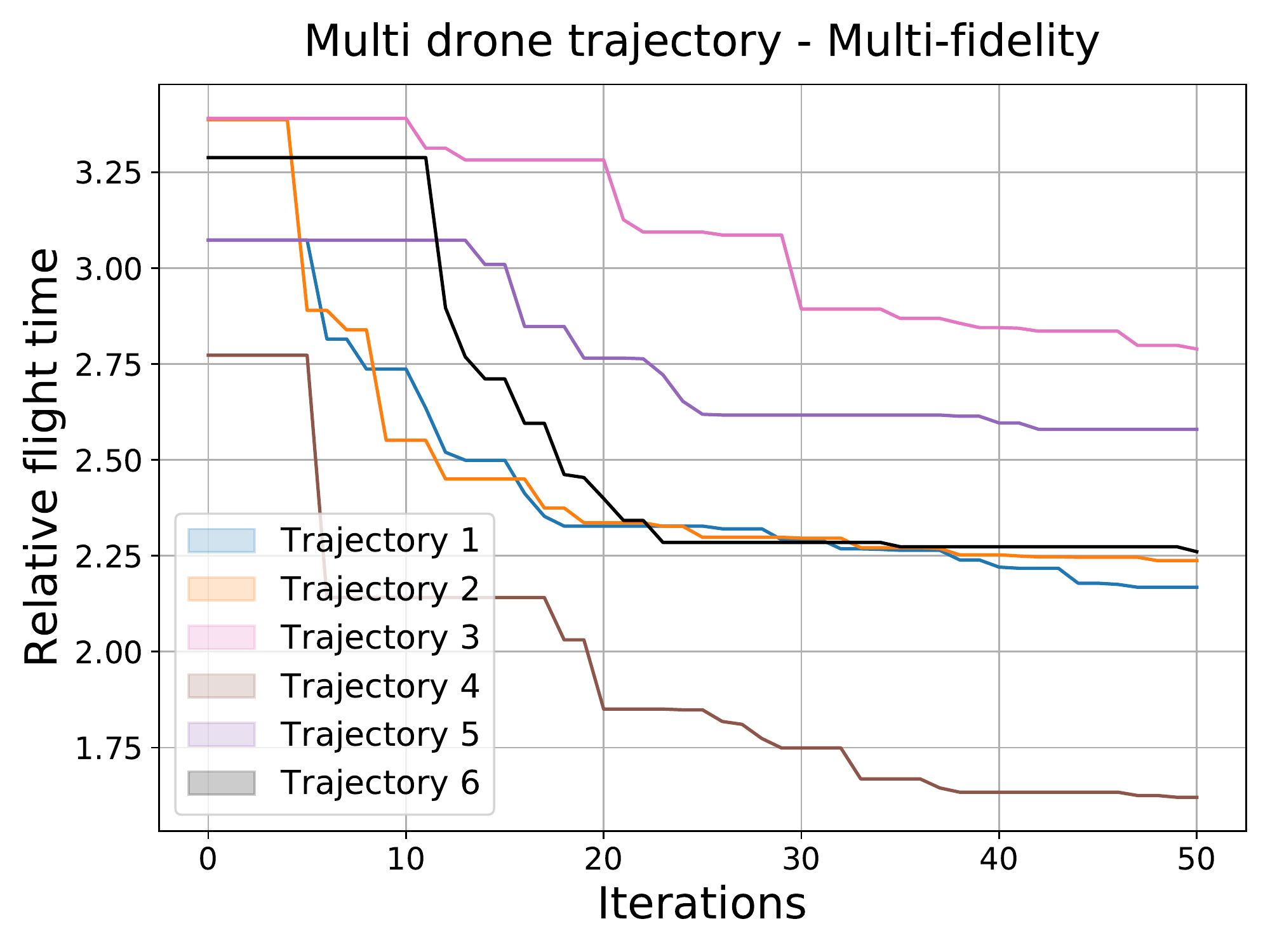}
	\caption{Relative flight time of multi-fidelity optimized trajectories compared to the initial best solution.}
	\label{fig:exp_sim_opt}
\end{figure}
\begin{figure*}
	\centering
    \includegraphics[width=\textwidth,trim=.0cm .0cm .0cm .0cm,clip]{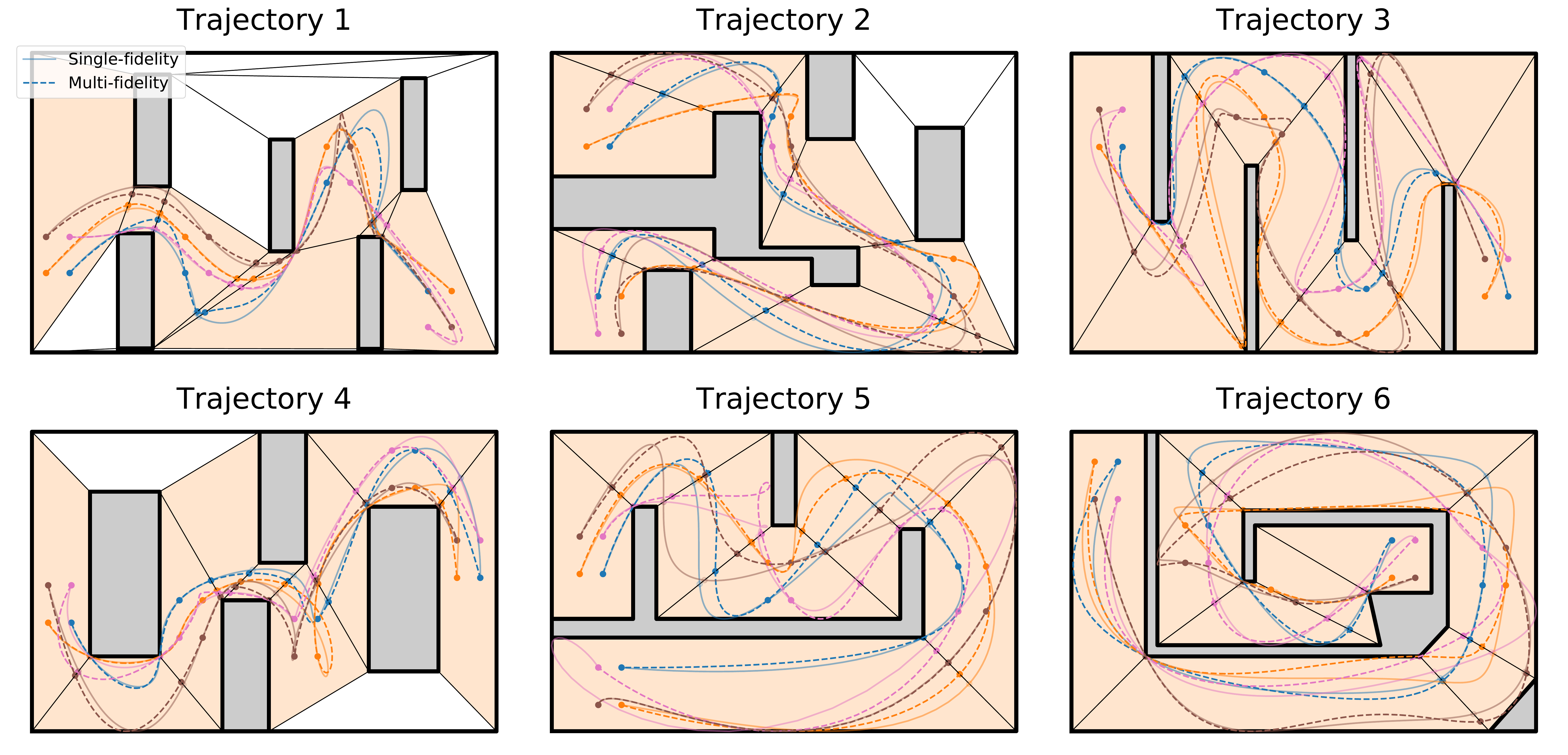}
    \captionsetup{width=0.9\textwidth, justification=centering}
	\caption{Single-fidelity (solid curve) and multi-fidelity (dashed curve) trajectory optimization results.}
	\label{fig:exp_trajectories_compare}
\end{figure*}

For the multi-fidelity experiments, the motor speed check based on differential flatness serves as low-fidelity evaluation and a more realistic 6-DOF flight dynamics simulation is used for high-fidelity evaluations.
We utilize the open-source multicopter dynamics and inertial measurement simulation by \cite{guerra2019flightgoggles} with the trajectory-tracking controller by \cite{tal2020accurate}.
For the high-fidelity evaluations, we set the maximum trajectory tracking error to 5 \si{cm} and the minimum distance between vehicles to 40 \si{cm}.
The initial best solution is obtained by slowing down the resulting trajectory from the single-fidelity optimization until the trajectory-tracking error bound is satisfied.
The low-fidelity dataset and the surrogate model are initialized with the dataset generated from the single-fidelity optimization.
We again use $h_{l_2} = 0.001$ and $\beta_{l_2} = 3.0$ for the acquisition functions \eqref{eqn:alg_acquisition_exploit_lower_bound} and \eqref{eqn:alg_acquisition_exploit}, but now set $C_{l_2} = 10$ to consider the increased evaluation cost.
For each trajectory, we run 50 iterations of Bayesian optimization.
At high-fidelity iterations at most four samples are evaluated, while low-fidelity iterations consider 64 samples.
Figure \ref{fig:exp_trajectories_compare} contains the resulting trajectories along with the trajectories obtained from the single-fidelity optimization.
The relative flight time for each iteration is presented in Fig. \ref{fig:exp_sim_opt}.

\begin{figure}
	\centering
	\includegraphics[width=0.35\textwidth,trim=.0cm .0cm .0cm .0cm,clip]{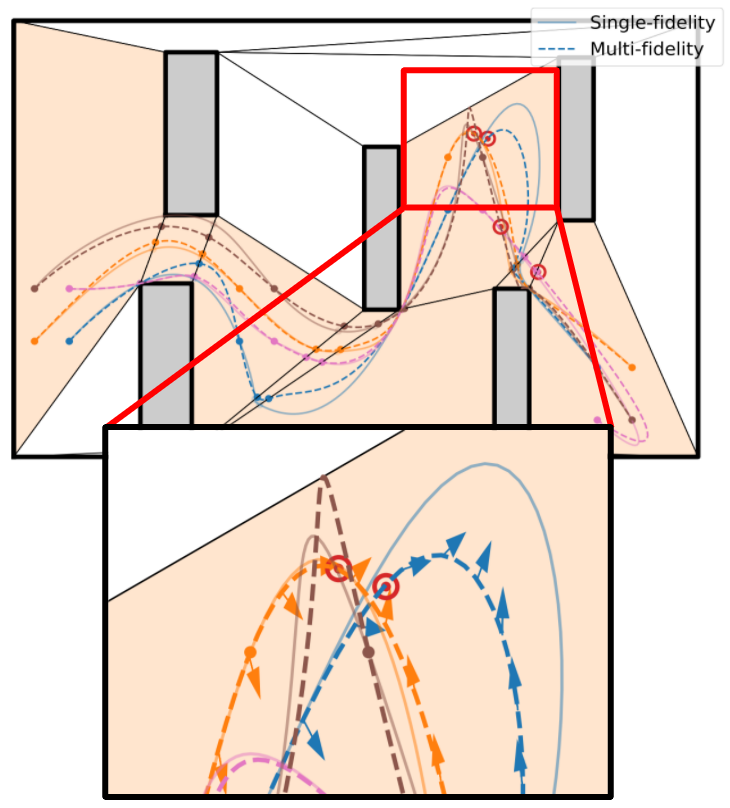}
	\caption{Single-fidelity (solid curve) and multi-fidelity (dashed curve) results for Trajectory 1. The red circles indicate the positions where the vehicles are closest to each other. The arrows indicate the tracking error, which is scaled 20 times. Since the direction of tracking error moves the blue trajectory away from the orange trajectory, the two trajectories can be placed closer together.}
	\label{fig:exp_sta_sim_compare}
\end{figure}

When examining Fig. \ref{fig:exp_trajectories_compare}, we observe that the multi-fidelity optimized trajectories significantly deviate from the single-fidelity trajectories.
Still, both sets of trajectories exhibit similar features.
For instance, in both cases, the orange vehicle in Trajectory 4 makes a loop to wait for the other vehicles.
Figure \ref{fig:exp_sta_sim_compare} provides an additional detailed comparison of part of Trajectory 1.
It can be seen that the single-fidelity trajectory of the blue vehicle takes a wide turn towards the top-right corner of the inset in order to avoid the orange vehicle.
In contrast, the corresponding multi-fidelity trajectory is much closer to the orange vehicle.
The multi-fidelity optimizer exploits the anticipated tracking error, which moves the blue vehicle away from the orange vehicle, to avoid the wide turn in the blue trajectory.

\subsection{Flight Experiments}

\begin{figure}
	\centering
	\includegraphics[width=0.46\textwidth,trim=.0cm .0cm .0cm .0cm,clip]{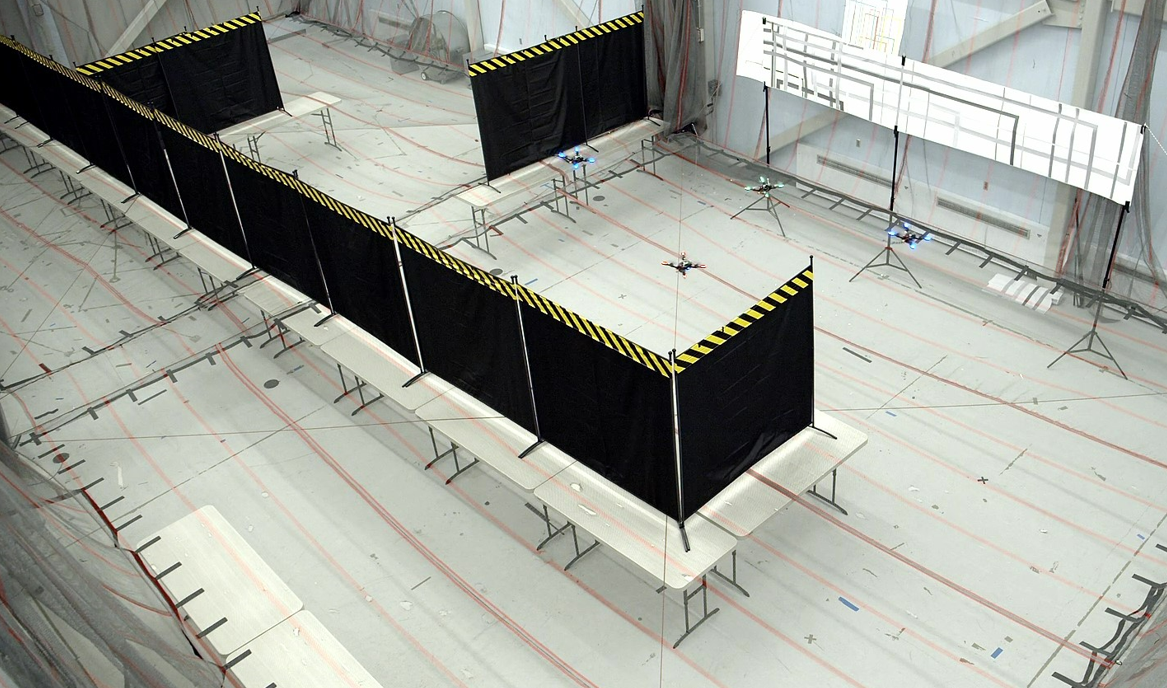}
	\caption{Environment for multi-agent flight experiments.}
	\label{fig:exp_real_drone_env}
\end{figure}

\begin{figure}
	\centering
	\includegraphics[width=0.48\textwidth,trim=.0cm .0cm .0cm .0cm,clip]{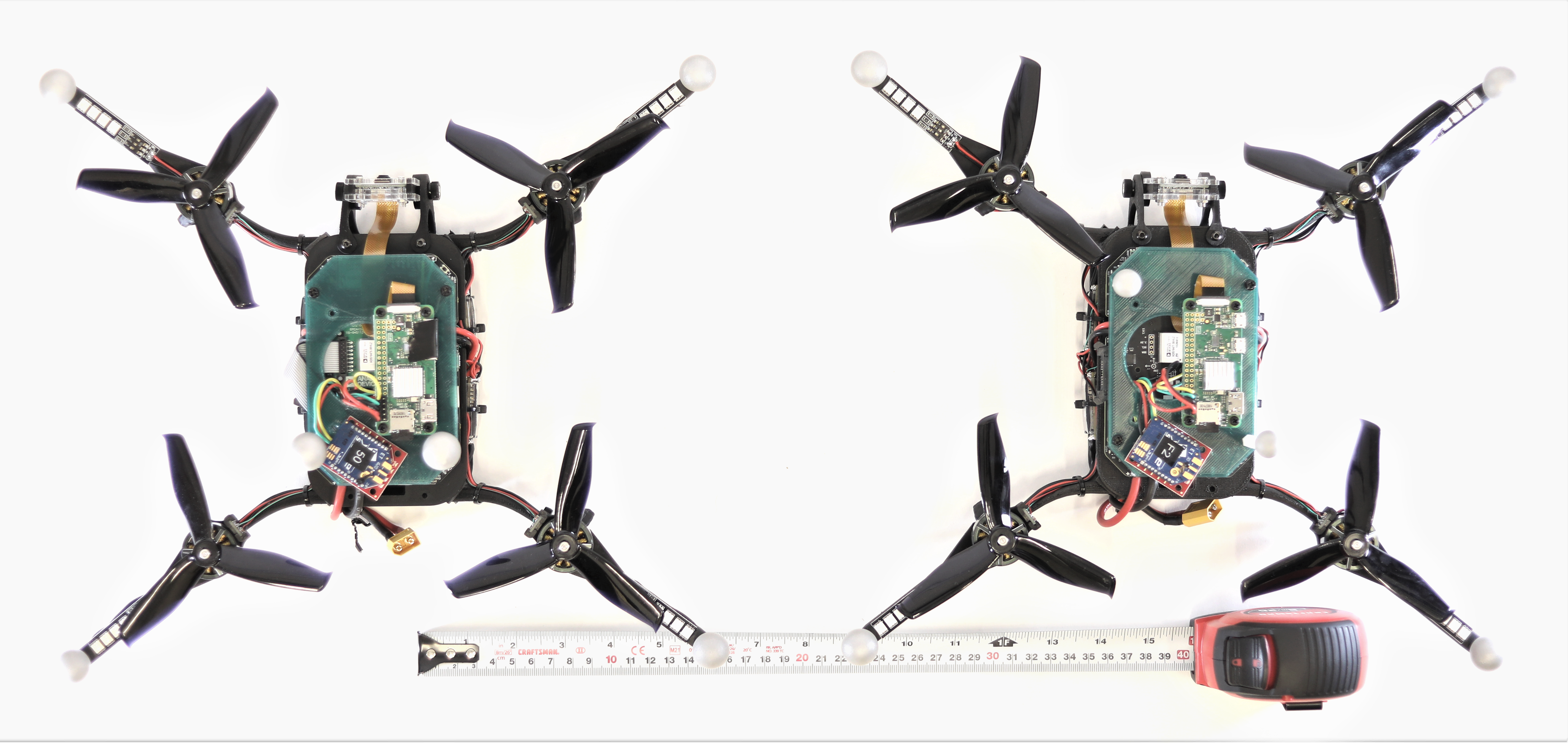}
	\caption{Quadcopter vehicles used for multi-agent flight experiments. Tape measure indicates 40 \si{cm}.}
	\label{fig:exp_real_drone}
\end{figure}

The trajectories obtained by the multi-fidelity optimization were further tested in real-world flight experiments in the motion capture space shown in Fig. \ref{fig:exp_real_drone_env}.
Figure \ref{fig:exp_real_drone} shows two of the four quadcopter vehicles used in the flight experiments, as well as a tape measure that illustrates their close proximity at the minimum distance of 40 \si{cm} between centroids.
The minimum separation between vehicles is not more than a few centimeters, while the trajectories reach speeds up to 7.4 \si{m/s}.

\begin{figure}
	\centering
	\includegraphics[width=0.35\textwidth,trim=.0cm .0cm .0cm .0cm,clip]{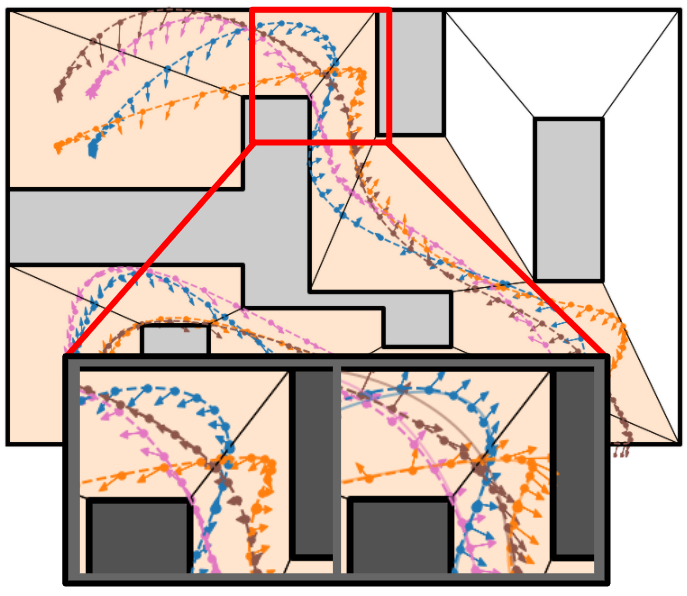}
	\caption{Tracking error in real-world flight for Trajectory 2. The arrows indicate the tracking error, which is scaled 20 times. The inset shows the actual (left) and anticipated (right) tracking error.}
	\label{fig:exp_sim_real_compare}
\end{figure}

The flight experiments demonstrate that the optimized trajectories can be flown on the real-world quadcopter vehicles.
A video of the experiments is included as supplementary material.
Due to the very small minimum distance, even a tiny discrepancy between the anticipated and actual trajectory-tracking error can result in a collision.
Indeed, we observed contact between two of the vehicles in Trajectory 1 and in Trajectory 2.
Despite the contact, all vehicles were able to continue flying and successfully completed every trajectory.
Figure \ref{fig:exp_sim_real_compare} shows the actual and anticipated tracking error around the point of contact in Trajectory 2.
It can be seen that the anticipated tracking error of the blue vehicle is in the wrong direction, resulting in contact with the brown vehicle.
The discrepancies between trajectory tracking in simulation and in real-world flight could potentially be addressed by incorporating actual flight experiments in the multi-fidelity optimization~\cite{ryou2021ijrr}.
\section{Conclusion} \label{sections:conclusion}
We have presented a novel modular Bayesian optimization algorithm to generate cooperative multi-agent trajectories subject to dynamic feasibility, obstacle avoidance, collision avoidance, and formation waypoint constraints.
The BayesOpt surrogate model is decomposed into multiple Gaussian process classifiers in order to alleviate the rapid increase in computational cost with increasing dimensionality.
Each module of the surrogate model approximates either individual dynamic feasibility constraints or pair-wise collision avoidance constraints.
By combining the acquisition function evaluation and the random sampling steps in BayesOpt, we further improve the overall efficiency of the Bayesian optimization.
The resulting algorithm is validated through simulation and real-world flight experiments in six unique environments with obstacles.

Future work may focus on resolving several limitations of the current algorithm.
Although we reduce the computation time with the modular structure and the efficient sampling method, the current implementation of our algorithm requires between two and three hours to optimize a multi-vehicle trajectory.
The main bottleneck is the updating time of the surrogate model, as the algorithm trains each module sequentially due to GPU memory limitations.
We expect that parallelization of the training process will reduce the computation time.

Another limitation of the algorithm is that it cannot check whether a feasible solution exists before running the optimization procedure. 
While the algorithm can find good solutions even when the initial trajectory is infeasible,
we have noticed that it often fails to find any feasible solution if there is a narrow corridor or---equivalently---if a large minimum distance between vehicles is required.
If a feasible solution exists, these cases could potentially be addressed by adjusting the training parameters.
This motivates us to consider a feasibility check that can establish the likely existence of a solution based on heuristic methods.
We conjecture that analyzing how the trajectory generation algorithm utilizes the shape of the polytopes may provide a way to estimate the existence of the solution.
Since the algorithm exploits the polytope shapes to shift or slow down the trajectory, the conditions for the optimal polygonal decomposition could inform whether a feasible solution exists.

\section*{Acknowledgments}
This work was partly supported by the Army Research Office through grant W911NF1910322.

\bibliographystyle{plainnat}
\bibliography{refs}

\end{document}